\def\year{2022}\relax
\documentclass[letterpaper]{article} 
\usepackage{aaai22}  
\usepackage{times}  
\usepackage{helvet}  
\usepackage{courier}  
\usepackage[hyphens]{url}  
\usepackage{graphicx} 
\usepackage{natbib}  
\usepackage{caption} 
\DeclareCaptionStyle{ruled}{labelfont=normalfont,labelsep=colon,strut=off} 
\usepackage[dvipsnames]{xcolor}         

\usepackage{pgfplots}
\pgfplotsset{compat=1.17} 
\usepackage{multirow}

\usepackage{algorithm}

\usepackage{newfloat}
\usepackage{listings}
\floatstyle{ruled}
\newfloat{listing}{tb}{lst}{}
\floatname{listing}{Listing}

\usepackage[utf8]{inputenc} 
\usepackage{booktabs}       
\usepackage{amsfonts}       

\usepackage{amsmath}
\usepackage{bm}
\usepackage{bbm}
\usepackage{todonotes}
\usepackage{algpseudocode}
\usepackage{subcaption}
\usepackage{svg}
\usepackage{amssymb}

\DeclareMathOperator*{\argmax}{arg\,max}
\DeclareMathOperator*{\argmin}{arg\,min}

\DeclareMathOperator*{\clamp}{clamp}

\newcommand\blfootnote[1]{%
  \begingroup
  \renewcommand\thefootnote{}\footnote{#1}%
  \addtocounter{footnote}{-1}%
  \endgroup
}

\title{Sparse Vicious Attacks on Graph Neural Networks}
\author{
   \author{
Giovanni Trappolini,\textsuperscript{1}\footnote{Corresponding Author at giovanni.trappolini@uniroma1.it}
Valentino Maiorca,\textsuperscript{2}$\dagger$
Silvio Severino,\textsuperscript{2}$\dagger$
Emanuele Rodolà,\textsuperscript{2}
Fabrizio Silvestri,\textsuperscript{1}
Gabriele Tolomei \textsuperscript{2}\\
\textsuperscript{1}{Department of Computer Engineering, Sapienza University of Rome}\\
\textsuperscript{2}{Department of Computer Science, Sapienza University of Rome}\\
}
}

\begin{document}

\maketitle

\newcommand{\Prob}{\mathbb{P}}
\newcommand{\R}{\mathbb{R}}
\newcommand{\Z}{\mathbb{Z}}
\newcommand{\E}{\mathbb{E}}
\newcommand{\insta}{\bm{x}}
\newcommand{\X}{X}
\newcommand{\G}{\mathcal{G}}
\newcommand{\advG}{\widetilde{\G}} 
\newcommand{\Gobs}{\G^{\text{obs}}}
\newcommand{\V}{\mathcal{V}}
\newcommand{\Vnew}{\V^{\text{new}}}
\newcommand{\Vadv}{\V^{\text{adv}}}
\newcommand{\edges}{\mathcal{E}}
\newcommand{\edgesnew}{\edges^{\text{new}}}
\newcommand{\edgesobs}{\edges^{\text{obs}}}
\newcommand{\edgesadv}{\edges^{\text{adv}}}
\newcommand{\graph}{\G=(\V,\edges)}
\newcommand{\neigh}{\mathcal{N}}
\newcommand{\adjM}{A}
\newcommand{\advadjM}{\widetilde{A}}
\newcommand{\adjMij}{{A}_{i,j}}
\newcommand{\dataset}{\mathcal{D}}
\newcommand{\train}{\dataset_{\text{train}}}
\newcommand{\test}{\dataset_{\text{test}}}
\newcommand{\features}{\mathcal{X}}
\newcommand{\labels}{\mathcal{Y}}
\newcommand{\hypspace}{\mathcal{H}}
\newcommand{\params}{\bm{\theta}}
\newcommand{\w}{\bm{\omega}}
\newcommand{\h}{\bm{h}}
\newcommand{\advh}{\widetilde{\bm{h}}}
\newcommand{\hyp}{h_{\params}}
\newcommand{\gnn}{g(\adjM, \X; \W)}
\newcommand{\model}{h^*}
\newcommand{\loss}{\ell}
\newcommand{\ladv}{\ell_{\text{adv}}}
\newcommand{\ldist}{\ell_{\text{dist}}}
\newcommand{\lnew}{\ell_{\text{new}}}
\newcommand{\Loss}{\mathcal{L}}
\newcommand{\savag}{\text{SAVAGE}}

\newcommand{\gabri}[1]{\todo[inline,color=BlueGreen!60]{{\bf Gabri:} #1}}
\newcommand{\fabri}[1]{\todo[inline,color=RedOrange!60]{{\bf Fabri:} #1}}
\newcommand{\ema}[1]{\todo[inline,color=ForestGreen!60]{{\bf Ema:} #1}}
\newcommand{\gio}[1]{\todo[inline,color=YellowOrange!60]{{\bf Giovanni:} #1}}
\newcommand{\silvio}[1]{\todo[inline,color=RoyalBlue!60]{{\bf Silvio:} #1}}
\newcommand{\vale}[1]{\todo[inline,color=Purple!60]{{\bf Valentino:} #1}}
\newcommand{\colored}[1]{\textcolor{red}{#1}}

\begin{abstract}
Graph Neural Networks (GNNs) have proven to be successful in several predictive modeling tasks for graph-structured data.
  Amongst those tasks, link prediction is one of the fundamental problems for many real-world applications, such as recommender systems.
  However, GNNs are not immune to adversarial attacks, i.e., carefully crafted malicious examples that are designed to fool the predictive model.
  In this work, we focus on a specific, white-box attack to GNN-based link prediction models, where a malicious node aims to appear in the list of recommended nodes for a given target victim. 
  To achieve this goal, the attacker node may also count on the cooperation of other existing peers that it directly controls, namely on the ability to inject a number of ``vicious'' nodes in the network.
  Specifically, all these malicious nodes can add new edges or remove existing ones, thereby perturbing the original graph.
  Thus, we propose \savag, a novel framework and a method to mount this type of link prediction attacks. 
  \savag\ formulates the adversary's goal as an optimization task, striking the balance between the effectiveness of the attack and the sparsity of malicious resources required.
  Extensive experiments conducted on real-world and synthetic datasets demonstrate that adversarial attacks implemented through \savag\ indeed achieve high attack success rate yet using a small amount of vicious nodes.
  Finally, despite those attacks require full knowledge of the target model, we show that they are successfully transferable to other black-box methods for link prediction.
\end{abstract}

\section{Introduction}
\label{sec:intro}

\blfootnote{$\dagger$ Equal Contribution}
The purpose of {\em link prediction} methods is to estimate the probability that an edge between two initially disconnected nodes in a graph should exist~\cite{martinez2016cs,kumar2020link}. 
For example, these methods are used to suggest the creation of new connections between social network users (e.g., follower-followee relationships).

Several approaches to link prediction have been proposed in the literature~\cite{lu2011link}. 
Recently, methods based on graph neural networks (GNNs) have achieved state-of-the-art performance~\cite{kipf2016semi}.

The list of predicted links for a given user ranked by their estimated probability represents a set of recommended new contacts, which eventually the user may decide to follow or not.
This step is crucial, especially for some authoritative, highly followed social network users, i.e., so-called ``{\em influencers}''.
Indeed, their reputation (and therefore social market value) may be affected by who they choose to follow.
Specifically, when the target of recommendations is an influencer, it is paramount to be accurate at suggesting the ``right'' set of candidate users to follow. 
As a matter of fact, being followed by an authoritative user can be seen as a sign of endorsement and ultimately could bring several advantages to the followee (e.g., higher revenue from advertising campaigns run on the social media platform due to increased popularity).
Hence, malicious users have a strong interest in {\em manipulating} link recommendation systems to inflate their reputation on social media platforms artificially, thus perpetuating their own (harmful) goals through adversarial attacks.
For example, consider Sammy is a social network user interested in increasing their reputation.
Ideally, they would have other influential users follow them to reach this goal..
However, Sammy has no control over other users, 
making such tactic unfeasible.
Still, Sammy could try to adjust their existing connections (i.e., follow/unfollow users) with the hope that their name will show up in the list of recommended connections for Terry, i.e., the user profile of a target influencer.
More generally, Sammy would control a subset of nodes (in addition to themselves) and is allowed to modify their neighborhood to achieve their goal. 
Sammy may also be able to inject {\em new} nodes in the network and create connections between such ``artificial'' users and other existing nodes.
Obviously, all these operations (e.g., adding/deleting an edge between two nodes) have a cost, which Sammy wants to account for and minimize. 
Although the body of work on adversarial attacks on link prediction is extensive~\cite{sun2018adversarial}, most of the literature do not completely cover the scenario described in the example above. 
Indeed, most work assumes the attacker can perturb the original network topology by adding or deleting connections only between {\em existing} nodes to favor the malicious goal~\cite{chen2020link}.
However, this setting is often unfeasible because it requires the direct control of real users, with prohibitive coordination costs.
To overcome this limitation, vicious node adversarial attacks have been proposed~\cite{wang2018attack}.
Under this setting, the attacker can create new users {\em ex-novo} and, by controlling these nodes, engineer the network to mount the attack.
This approach is much more feasible than directly controlling real nodes; indeed, the Internet is full of services selling fake accounts/comments/likes.


Still, existing methods for implementing vicious node attacks do not incorporate the cost of adding such malicious nodes into the adversary's objective~\cite{wang2018attack}. 
Those methods typically assume the existence of an upper bound on the attacker's budget allocation, which is generally exhausted to favor the highest attack success rate possible.
In other words, if an attacker has a maximum allowed budget of $N$ vicious nodes to inject into the network, it will use all of them to increase the chance of attack success.
We argue that this solution is inefficient and impractical.
For this reason, in this paper we introduce \savag\ ({\em \textbf{S}p\textbf{A}rse \textbf{V}icious \textbf{A}ttacks on \textbf{G}raph n\textbf{E}tworks}).
This is a novel framework that provides an original perspective on vicious node adversarial attacks on GNN-based link prediction methods for {\em directed} graphs. 
\savag\ operates in a white-box setting and promotes a spare use of the malicious resources required by the attacker, i.e., the number of vicious nodes and their aggressiveness.
Moreover, \savag\ is general enough as it allows to frame existing vicious node attacks into it. 

\begin{figure*}
    \centering
    \begin{subfigure}[h]{0.3\textwidth}
        \includegraphics[width=\textwidth]{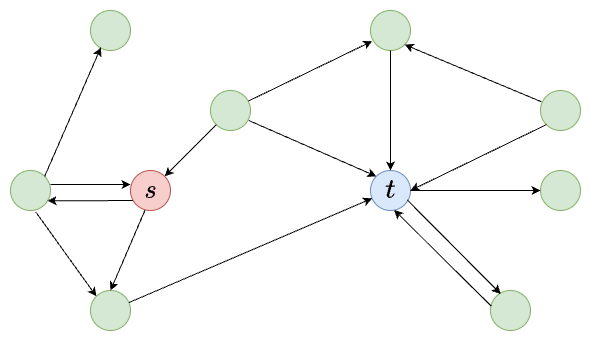}
        \caption{The original {\em directed} graph.}
        \label{fig:graph-ori}
    \end{subfigure}
    ~ 
    \begin{subfigure}[h]{0.3\textwidth}
        \includegraphics[width=\textwidth]{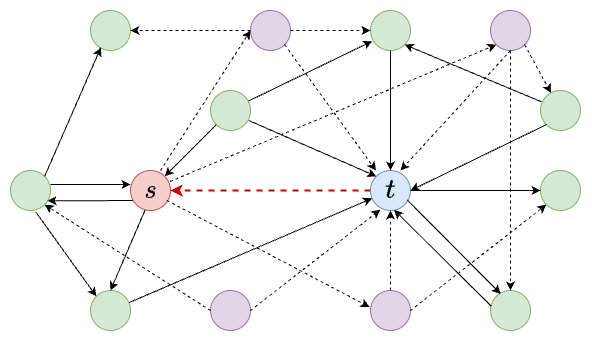}
        \caption{A generic attack to mispredict the existence of the link between $t$ and $s$.}
        \label{fig:graph-perturbed}
    \end{subfigure}
    ~ 
    \begin{subfigure}[h]{0.3\textwidth}
        \includegraphics[width=\textwidth]{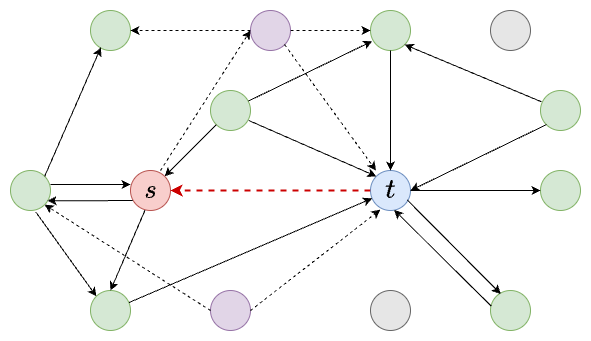}
        \caption{The attack generated by \savag\ is still successful but uses a sparser number of malicious resources.}
        \label{fig:graph-perturbed-savage}
    \end{subfigure}
    \caption{The picture on the left (a) shows a simple directed graph, where the red node indicates the {\em source} ($s$) and the blue node the {\em target} ($t$) of the link prediction attack. In the middle (b), a generic attack is depicted: the injection of vicious nodes (in purple) and new connections induce the link prediction system to suggest the connection between $t$ and $s$. Typically, these attacks exhaust the allocated attacker's budget. On the other hand, \savag\ (c) enforces sparsity on the amount of malicious resources used, deactivating unnecessary vicious nodes (grayed-out), while keeping the attack successful.}\label{fig:teaser}
\end{figure*}

    

Overall, the main novel contributions of this paper are as follows:
\textbf{(a)} We introduce \savag, a novel framework that enables a comprehensive attack model based on vicious nodes to generate adversarial attacks on GNN-based link prediction systems;
\textbf{(b)} \savag\ trades off the power of the attack with the number of malicious resources required by explicitly enforcing the sparsity of vicious controlled nodes without assuming any constraint on the attacker's budget;
\textbf{(c)} We show the feasibility and effectiveness of \savag\ in attacking several GNN-based link prediction models learned from real-world and synthetic datasets through extensive experiments;
\textbf{(d)} We demonstrate that adversarial attacks generated with \savag\ can be successfully transferred to harm other link prediction systems (not necessarily GNN-based) under the more challenging black-box setting.

All code and data is released as supplementary material.


\section{Related Work}
\label{sec:related} 
\paragraph{Link Prediction.}
Link prediction is one of the most investigated problems in modern graph analysis.
One class of simple yet effective approaches for link prediction is called heuristic methods.
These assume that the existence of a link between two nodes depends on their ``similarity''. 
In practice, they use some predefined heuristics to compute node similarity scores as the link probability~\cite{liben2003cikm,lu2011link}.
Popular heuristics include: Common Neighbors~\cite{newman2001clustering}, Jaccard~\cite{liben2003cikm}, Preferential Attachment~\cite{barabasi1999pa}, Adamic-Adar~\cite{adamic2003sn}, Resource Allocation~\cite{zhou2009predicting}, Katz~\cite{katz1953new}, PageRank~\cite{brin1998pagerank}, and SimRank~\cite{jeh2002simrank}.

Although working well in practice, heuristic methods have strong ``hand-coded'' assumptions on when links may exist.
To overcome this limitation, the link prediction problem has been formulated as a standard binary classification task and solved using well-known supervised learning techniques~\cite{alhasan2006sdm}.

With the advancements of deep learning and, specifically, graph neural networks (GNNs), several approaches have recently proposed effective GNN-based link prediction methods~\cite{zhang2018neurips,li2021icml}.
Roughly speaking, GNNs allow to learn suitable representations (i.e., embeddings) of graph nodes by aggregating information derived from the local neighborhood of each individual node.
These node embeddings are in turn used as input of a downstream link prediction function, thus making the whole model's architecture end-to-end trainable. 
In this work, we focus on a specific attack to GNN-based link prediction.

For a comprehensive survey on link prediction, we refer the reader to~\cite{martinez2016cs,kumar2020link,wu2020comprehensive}.



\paragraph{Adversarial Attacks to Link Prediction.}
Although very powerful, studies have shown that machine learning models may be vulnerable to so-called {\em adversarial attacks}, i.e., carefully crafted malicious examples designed to fool the predictive models. 
Typically, these adversarial inputs are generated by introducing minor -- yet thoughtfully selected -- perturbations to regular inputs. 
These attacks have been widely proven successful in many critical domains, such as image recognition~\cite{goodfellow2014explaining} and malware detection~\cite{grosse2017esorics}.
However, a few works have explored how effective such adversarial attacks may be for link prediction algorithms, especially those based on GNNs, which require the attacker to perturb the input graph. 
Amongst these studies, it is worth mentioning~\cite{zugner2018adversarial, chen2020link,lin2020adversarial}. We refer the reader to~\cite{dai2022comprehensive} for a comprehensive survey.

The methods above assume that the attacker can control a subset of existing nodes. As stated in Section~\ref{sec:intro}, though, this is a pretty unrealistic premise, as it would be prohibitively expensive. 
To address this issue,~\cite{wang2018attack} introduce the capability for the attacker to generate {\em new} (i.e., fake/vicious) nodes to mount the attack more efficiently.
Furthermore,~\cite{wang2020scalable} propose a linear approximation of the previous method to make it more scalable. 
\citet{dai2022targeted} formulate a universal perturbation that can target multiple nodes and still remain effective. 

In both classic and vicious settings, all the methods presented so far introduce an unnoticeability constraint of the graph perturbation. This constraint is usually of two kinds: either the attacker is given a fixed budget to spend~\cite{chen2020link}, or some rules are imposed to control the difference between the original and the perturbed graph, like in~\cite{lin2020adversarial}.
Either way, this is treated as an upper bound on the amount of ``malicious resources'' used to implement the attack and does not promote frugality. 
In fact, the adversary typically saturates the constraint to ensure that the attack is successful.
On the other hand, the method we propose in this work enforces the sparsity of the malicious resources used by the attacker (see Fig.~\ref{fig:teaser}).

\section{Background and Notation}
\label{sec:background}
We consider a {\em directed} graph $\graph$ with $n$ nodes $\V$, and $m$ edges $\edges$.
The structure of $\G$ is encoded by its adjacency matrix $\adjM \in \lbrace 0,1 \rbrace^{n \times n}$, where $\adjMij$ = 1 iff $(i, j) \in \edges$. 
Notice that, in general, $\adjMij \neq A_{j,i}$, i.e., the adjacency matrix is not symmetric.
A feature matrix $\X \in \mathbb{R}^{n \times k}$ can also be used to associate features to nodes of $\G$.
We assume that a graph neural network (GNN) $g$ learns a hidden representation of nodes in the graph (i.e., a node embedding). Such representation is, in turn, used for our downstream task of interest (i.e., link prediction).

The embedding of each node is learned through $g$ by iteratively updating the node's features based on the neighbors' features.
Formally, let $\h^l_u$ denote the embedding of node $u\in \V$ at the $l$-th layer of $g$. 

Let $\neigh^1(u) = \neigh(u) = \{v\in \V~|~(u,v)\in \edges\}$ be the $1$-hop neighborhood of $u\in \V$, i.e., the set of all nodes that are adjacent to $u$.
More generally, we can define the $l$-hop neighborhood of $u$ ($l\in \Z^+,~l > 1$) as the set of nodes that are {\em at most} $l$ hops away from $u$, using the following recurrence relation:
\[
\neigh^l(u) = \neigh^1(u) \cup \bigcup_{v\in \neigh(u)} \neigh^{l-1}(v).
\]
Hence, let us consider the subgraph $\G^l_u$ of $\G$ induced by $u$ and its $l$-hop neighborhood $\neigh^l(u)$ as relevant for the computation of $\h^l_u$. 
Specifically, we consider $\adjM^l_u$ and $\X^l_u$ as the adjacency matrix and the feature matrix of nodes in the subgraph $\G^l_u$, respectively, and $\h^l_u$ is computed by $g$ using the following equation:
\begin{equation}
    \label{eq:gnn-emb}
    \h^l_u = g(\adjM^l_u,\X^l_u;\params_g) = \phi(\h^{l-1}_u, \psi(\{\h^{l-1}_v~|~v\in \neigh(u)\})),
\end{equation}
where $\phi$ and $\psi$ are arbitrary differentiable functions (i.e., neural networks): $\psi$ is a permutation-invariant operator that {\em aggregates} the information from the $l$-hop neighborhood of $u$; $\phi$ {\em updates} the node embedding of $u$ by combining information from the previous layer; $\params_g$ are the trainable parameters of $g$.
The number of hidden layers in $g$ determines the set of neighbors included while learning each node's embedding.
From now on, unless otherwise needed, we will consider $l$ fixed and omit the corresponding superscript.

Given the graph $\Gobs=(\V, \edgesobs)$, where $\edgesobs \subset \edges$ is a subset of the true links observed, the {\em link prediction} problem generally resorts to estimating the probability that an edge exists between two nodes $u$ and $v$, i.e., $\Prob[(u,v)\in \edges \setminus \edgesobs]$.

Let $\h_w = g(\adjM_w,\X_w;\params_g)$ be the embedding of the generic node $w\in \V$.
Thus, we assume that the predicted link probability between two nodes $u$ and $v$ is approximated with a function $f$, defined as follows:
\begin{equation}
\label{eq:gnn-lp}
    f(\h_u, \h_v; \params_f) = f (g(\adjM_u,\X_u;\params_g), g(\adjM_v,\X_v;\params_g); \params_f).
\end{equation}

Notice that $f$ can itself be another neural network (e.g., an MLP) that takes as input two node embeddings learned by the GNN $g$ and outputs their link prediction.


A combined embedding $\h_{u,v}$ of the two input nodes can be obtained, for example, via Hadamard (element-wise) product, i.e., $\h_{u,v} = \h_u \odot \h_v$.
Eventually, the probability of a link existing between $u$ and $v$ can be computed as $f(\h_u, \h_v;\params_f) = \sigma(\params_f^T\h_{u,v})$, where  $\sigma(x) = \frac{1}{1+e^{-x}}$ is the sigmoid activation function at the last output layer of $f$.

In the following, we denote as $\params$ the overall set of end-to-end trainable parameters of $f$ and $g$, i.e., $\params = \left[\params_f, \params_g\right]$.

\section{Problem Formulation}
\label{sec:problem}

\subsection{Attack Model}
\label{subsec:attack-model}
We consider a snapshot of a network, modeled as a {\em directed} graph $\graph$, and a trained GNN-based link prediction model $f$. 
In our attack model, we assume the existence of an attacker node ($s\in \V$) that can perturb the original graph $\G$ into $\advG = (\V', \edges')$.
The perturbation represents the capability of the attacker to add a set of new vicious nodes $\Vnew$ to $\G$, such that $\V'=\V \cup \Vnew$. Furthermore, we suppose the attacker controls a subset of the original nodes $\V^s \subseteq \V$ (trivially, $s\in \V^s)$ and is capable of adding/removing edges starting from the set of governed nodes $\V^s \cup \Vnew$.
Let $\edges^+ \subseteq (\V^s \cup \Vnew)\times \V'$
and $\edges^- \subseteq (\V^s \cup \Vnew)\times \V'$ be the set of edges added and removed by the attacker, respectively.
Eventually, $\advG = (\V', \edges')$ denotes the final perturbed graph, where $\edges' = \edges \cup \edges^+ \setminus \edges^-$.
Thus, the goal of the attacker $s$ is to transform $\G$ into $\advG$, using the capabilities described above, to ultimately induce the link prediction model $f$ in recommending $s$ to a target victim node $t \in \V \setminus (\V^s \cup \Vnew)$, i.e., to suggest the creation of the directed edge $(t,s)$ between the victim and the attacker.
We also assume the attacker has full knowledge of $f$, i.e., we consider a {\em white-box} threat model, where the architecture and the entire set of parameters of $f$ ($\params = \left[ \params_f, \params_g \right ]$) are known and fixed~\cite{pitropakis2019csr,ren2020eng}.
Later in Section~\ref{subsec:transf}, we show that this requirement does not limit the adversary's power, as our proposed attack can be successfully transferred to disrupt other link prediction models in a {\em black-box} setting.

\subsection{Attack Framework}
\label{subsec:attack-framework}
The attacker $s$ aims to maximize the chance that the directed edge between the target victim $t$ and itself, i.e., $(t,s)$, will be amongst those predicted by $f$ while minimizing the ``malicious effort'' needed.

More formally, suppose $f$ is defined as in Eq.~\ref{eq:gnn-lp}.
To ease the notation, we will omit the parameters $\params_f$ and $\params_g$ from the equations below.
Also, let $\adjM$, $\adjM'$, and $\advadjM$ be the adjacency matrices associated with the original graph $\graph$, the intermediate graph $\G' = (\V', \edges)$, and the final perturbed graph $\advG = (\V', \edges')$, respectively. 
Thus, the problem for the attacker $s$ is to find the optimal perturbation $\advh^*_s \neq \h_s$ such that $f(\h_t, \advh^*_s)\neq f(\h_t, \h_s)$ by solving the objective below:
\begin{equation}
\label{eq:optimal-emb}
\begin{aligned}
\advh^*_s~=~& \argmax_{\advh_s} \Big\{ \big[f(\h_t, \advh_s) - f(\h_t, \h_s)\big] - d(\advh_s, \h_s) \Big\}, \\
\end{aligned}
\end{equation}
where $f(\h_t, \advh_s) - f(\h_t, \h_s)$ measures the difference between the original and the adversarial link prediction, and $d$ captures the magnitude of the malicious effort used by the attacker for transforming $\h_s$ to $\advh_s$.

The generic perturbation $\advh_s = g(\advadjM_s, \X_s)$ indicates the attacker node embedding output by a trained GNN $g$, after the $1$-hop neighborhood of the nodes controlled by the attacker $\V^s \cup \Vnew$ has been modified according to its capabilities (i.e., by adding/removing edges as specified by $\edges^+$ and $\edges^-$, respectively).
Specifically, we assume there exists a function $\pi$ that works as follows:
\begin{equation}
\label{eq:pi}
    \advh_s = g(\pi(\adjM'), X_s) = g(\advadjM_s, \X_s),
\end{equation}
where $\adjM'$ is the full adjacency matrix associated with the intermediate graph $\G'$, and $\advadjM_s = \pi(\adjM')$.
Notice that the perturbation applied by $\pi$ reflects the attacker's capabilities described in our attack model, i.e., it concerns {\em only} the topological structure of the neighborhood of the nodes controlled by the attacker, whereas the node feature matrices are left unaltered. 
We acknowledge that the ability to modify also the node feature matrices is an interesting direction to explore, and we plan to investigate this as future work since our framework is general enough to cover also that scenario.
Moreover, the perturbation induced by $\pi$ on $\h_s$ must not affect the embedding of the target victim node $\h_t$.
According to Eq.~\ref{eq:gnn-emb}, the hidden representation of a node $u$ produced by a given $l$-layer GNN ($\h_u$) is influenced by all the other nodes $v$ that are in the $l$-hop neighborhood of $u$, i.e., $\neigh^l(u)$. 
Thus, to avoid that $\h_t$ is affected by $\pi$, we assume $t\notin \neigh^l(s)$ and, even strongly, $s\notin \neigh^l(t)$.

To measure the malicious effort of the attacker, the function $d$ captures the difference between the original and the perturbed attacker node embeddings.
Such a difference must consider: {\em(i)} the number of {\em additional} vicious nodes $\Vnew$ injected by the attacker to generate the intermediate augmented graph $\G'$; {\em(ii)} the distance between the original adjacency matrix $\adjM_s$ and the perturbed $\advadjM_s$ obtained from structural modifications (i.e., edge additions/deletions) of $\G'$. 

\section{Method}
\label{sec:method}
In this section, we describe the method used by the attacker to solve the optimization problem defined in Eq.~\ref{eq:optimal-emb}, which we overall refer to as \savag.

We consider the function $\pi$ parametrized by a perturbation matrix $P$, i.e., $\pi(\cdot;P)$, and thus rewrite Eq.~\ref{eq:pi} as follows:
\begin{equation}
\label{eq:pi_v2}
    \advh_{s;P} = g(\pi(\adjM'; P), X_s) = g((P\oplus \adjM'), X_s) = g(\advadjM_s, \X_s),
\end{equation}
namely $\advadjM_s = \pi(\adjM'; P) = P\oplus \adjM'$, where $\oplus$ is the element-wise matrix addition.

The perturbation matrix $P\in \{-1,0,1\}^{n\times n}$ is a discrete squared matrix, where $n = |\V'|$ is the total number of nodes of the augmented graph $\G' = (\V',\edges)$. 
This includes the set of additional vicious nodes $\Vnew$ injected by the attacker into the original graph $\graph$ (i.e., $\V' = \V\cup \Vnew$), yet keeping the same set of original edges $\edges$, thus is represented by the adjacency matrix $\adjM'$. 
For each $i\in \V^s \cup \Vnew, j \in \V'$:
\[
  P_{i,j} =
  \begin{cases}
    +1 & \text{if $(i,j)\in \edges^+$: {\em add} the edge between $i$ and $j$,}\\
    -1 & \text{if $(i,j)\in \edges^-$: {\em remove} the edge between $i$ and $j$,}\\
    0 & \text{otherwise.}
  \end{cases}
\]
Hence, for a given perturbation matrix $P$, a graph $\G'$ with $\Vnew$ vicious nodes, a pair of source (attacker) and target (victim) nodes $s,t\in \V$, and a fixed GNN-based link prediction model $f$, such that $(t,s)\notin \edges \wedge f(\h_t,\h_s)=0$, we can compute the following (instance-level) loss function:
\begin{equation}
    \label{eq:instance-loss}
    \Loss(P) = \ladv[f(\h_t, \advh_{s;P})] + \beta \ldist(\h_s, \advh_{s;P}) + \gamma \lnew(\bm{v}^{\text{new}})
\end{equation}
The first component ($\ladv$) penalizes when the adversarial prediction goal is {\em not} satisfied and can be computed as:
\begin{equation}
\label{eq:lpred2}
\ladv[f(\h_t, \advh_{s;P})] = - \log[f(\h_t, \advh_{s;P})],
\end{equation}
which corresponds to a standard binary cross-entropy, where the class label to predict is always $1$, i.e., we want to enforce the prediction of an adversarial edge between $t$ and $s$.

The second component ($\ldist$) is 
an arbitrary distance function that discourages $\advh_{s;P}$ from being too far away from $\h_s$, namely $\advadjM_s$ must be close to the original $\adjM_s$.
For example, $\ldist(\h_s, \advh_{s;P}) = ||\adjM_s - \advadjM_s||_p$, where $||\cdot||_p$ is the $L^p$-norm. 

The third component ($\lnew$) controls the number of vicious nodes injected into the original graph network.
In practice, $\bm{v}^{\text{new}}$ is a $|\Vnew|$-dimensional binary vector (i.e., a mask) indicating the vicious nodes injected by the attacker, i.e., $\bm{v}^{\text{new}}[u] = 1$ iff node $u \in \Vnew$ has been added to the graph network, $0$ otherwise.
For example, $\lnew(\bm{v}^{\text{new}}) = ||\bm{v}^{\text{new}}||_p$. 
In this work, we set $p=1$ both for $\ldist$ and $\lnew$.

Eventually, solving the objective defined in Eq.~\ref{eq:optimal-emb} is framed into the following optimization task:
\begin{equation}
    \label{eq:opt-adv-lp}
    \advh_{s;P^*}=\argmin_{P} \Loss(P).
\end{equation}
It is worth noticing that, in this original formulation, the optimization objective defined in Eq.~\ref{eq:opt-adv-lp} is inherently discrete. 
To make it smooth, inspired by prior work~\cite{lucic2022cf,srinivas2017cvprw}, we first consider an intermediate, real-valued perturbation matrix $\hat{P}$ with entries in $(-1,1)$, obtained by applying a hyperbolic tangent transformation ($\tanh$).
Intuitively, this matrix indicates the degree of confidence of the attacker to add or remove an edge from the adjacency matrix $\adjM'$.
Thus, we can replace $P$ with $\hat{P}$ in Eq.~\ref{eq:opt-adv-lp}, and solve the objective via standard gradient-based optimization methods like stochastic gradient descent or similar.
Eventually, to obtain the discrete matrix $P$, we simply threshold the entries of $\hat{P}$ as follows:
\[
  P_{i,j} =
  \begin{cases}
    +1 & \text{if $\hat{P}_{i,j} \geq t^+$,}\\
    -1 & \text{if $\hat{P}_{i,j} \leq t^-$,}\\
    0 & \text{otherwise.}
  \end{cases}
\]
A straightforward choice for the thresholds above is: $t^+ = 0.5$ and $t^- = -0.5$.
Notice, though, that $P\oplus A'$ would lead to a discrete matrix $\advadjM_s$ whose entries are in the set $\{-1,0,1,2\}$, instead of $\{0,1\}$ as required.
Indeed, removing an edge that does not exist from $\edges$, will lead to an entry equals to $-1$; on the other hand, adding an edge that already exists in $\edges$ will result in an entry equals to $2$.

We therefore obtain the correct final perturbed matrix $\advadjM_s$ by applying a $\clamp_{[0,1]}$ function element-wise, i.e., $\advadjM_s[i,j] = \clamp_{[0,1]}(P_{i,j} + A'_{i,j})~\forall (i,j)$, where $\clamp_{[0,1]}(x) = \max(0,\min(x,1))$.

\section{Experiments}
\label{sec:experiments}

We evaluate the effectiveness of \savag\ on several challenges.
First, we assess the power of our attack model by reporting some key performance metrics.
Second, we analyze the impact of the critical components of our model through ablation studies.
Finally, we demonstrate the ability of \savag\ to generate attacks that transfer toward other black-box link prediction systems.
We have released code on Github \footnote{https://github.com/GiovanniTRA/SAVAGE}.

\subsection{Experimental Setting}

\paragraph{Datasets.}
We validate our method both on real-world and synthetic datasets.
Moreover, as dictated by our attack model, we focus only on data representing {\em directed} graphs. 
Our real world datasets are mainly of two kinds: citation-based and social networks.
The former category includes \emph{Cora}~\cite{kunegis2013konect, vsubelj2013model}, \emph{Arxiv}~\cite{wang2020microsoft}, and \emph{Citation2}~\cite{wang2020microsoft}, whereas the latter contains \emph{Twitter}~\cite{leskovec2012learning}, \emph{Wikipedia-Vote}~\cite{leskovec2010signed, leskovec2010predicting}, and \emph{GPlus}~\cite{leskovec2012learning}.
Furthermore, we test our method on a synthetic dataset, partially taken from~\cite{Fey2019FastGR}, that we call \emph{Synthetic}.
It is worth noticing that, to properly test our method, we choose datasets featuring a varying degree of cardinality and density.
For further details on the preprocessing steps taken and additional statistics, we refer the reader to the supplementary material.

\paragraph{GNN Models.}
To evaluate the effectiveness of \savag, we first train a GNN-based link prediction system. The model is composed of two stacked convolutional layers of 128 and 64 hidden units, respectively, and a two-layer MLP with a sigmoid activation function at the last output layer.
In our setup, we minimize the prediction error with a binary cross-entropy loss using Adam optimizer with $3\times 10^{-3}$ learning rate. 
Finally, we compute the accuracy (using $0.6$ as classification threshold) and the AUROC (Area Under the ROC) curve of the learned GNN-based link predictor.  
On average, the best model achieves $85\%$ accuracy and $0.86$ AUROC score on every dataset considered. 
The full details are available in the supplementary material.

\paragraph{Methods.}
We test our method against different competitors and under several settings.
First, we consider a simple random baseline (RAND). 
This can add/remove connections in the graph and activate/deactivate vicious nodes with a probability $p$; the larger $p$, the stronger the corresponding attacker. 
We denote by RAND-L and RAND-H the baseline with the lowest and the highest probability of addition/activation, i.e., $p_\text{L} = 0.25$ and $p_\text{H} = 0.75$.
While we cannot perform a direct comparison with other methods, as they are not naturally designed for our setting, we adapt the greedy technique proposed in~\cite{wang2018attack} and~\cite{chen2020link} to work in our framework, and we call it AIGA (Adapted Iterative Gradient Attack). 
Finally, we consider four instances of our method: {\em(i)} \savag, {\em(ii)} \savag-N,  {\em(iii)} \savag-I, and  {\em(iv)} \savag-NI. 
\savag\ refers to the plain full framework, randomly initialized and defined in Eq.~\ref{eq:instance-loss}; \savag-N is the same as \savag\ yet without the penalty losses (i.e., $\ldist$ and $\lnew$); \savag-I is the same as \savag\ yet initialized with the output from AIGA; and \savag-NI is the same as \savag-N yet initialized with the output from AIGA.
Since \savag\ tries to solve a highly non-convex problem, the initialization of the parameters is indeed crucial. Distinguishing between random and AIGA-based initialization serves two purposes.
First, it shows that \savag\ can be used just as a sparsifying framework, compatible with other techniques; second, it demonstrates the impact of initialization on the optimization problem.
For all the models considered, performance is reported according to their best hyperparameter setting, which is detailed in the supplementary material.

\paragraph{Evaluation Metrics.}
Given a graph and a GNN-based link prediction model trained on it, we uniformly sample a source and a target node from the graph, providing that the two nodes are disconnected and, even further, they are not part of each other's 2-hop neighborhood. Moreover, according to the link prediction model, no edge should exist between the two sampled nodes. 
Then, we attack the graph using all the methods described above (i.e., the four variants of our \savag\ along with its competitors) and run again the link prediction on the resulting perturbed graph.
We repeat this procedure for 20 different node pairs, for each dataset used.
We present our results according to the following metrics.
The average prediction under attack (AP) is the estimated probability output by the link prediction model {\em after} the attack.
The attack rate (AR) measures the fraction of successful attacks, i.e., it counts the fraction of predictions above the success threshold ($0.6$). 
AP and AR capture two different aspects of the attack power. The former indicates the average {\em strength} of each attack (i.e., the new link probability between the source and target node, which initially was below the success threshold). The latter measures only how frequently the new link probability is large enough for the attack to succeed, regardless of its actual value.
Finally, we report statistics on the perturbation induced by \savag\ on the original graph topology: the number of vicious nodes injected (AN) and the shift in the node's degree distribution {\em before} and {\em after} the attack as measured with the Kullback-Leibler divergence (KL).


\subsection{Attack Power}
We measure the power of the attacks generated by all the methods examined on the real-world and synthetic datasets; the results are shown in Table~\ref{tab:vn_result}.
We may observe that \savag\ can mount very effective attacks across all datasets, with the highest AP and AR between all the considered methods. 
In terms of AR, we notice a positive correlation between the performance gap and the density of the dataset.
Furthermore, \savag\ significantly reduces the number of malicious resources used (AN) w.r.t. to AIGA, ranging from a minimum of 20\% up to 80\% decrease and the KL score illustrates the limited impact on graph connectivity, especially when compared with the baselines (see Section~\ref{subsec:limit} for additional discussion). 
We observe that the AIGA initialization can further mitigate this impact at the cost of a slightly worse AR in sparser graphs. This may be caused by the initialization that imposes the solution to get stuck around the starting local minimum.
In other words, this analysis shows that \savag\: {\em (i)} generates more successful attacks than competitors, {\em (ii)} with stronger confidence, and {\em(iii)} does not rely on the introduction of random noise since random baselines are not able to disrupt the prediction even when using a comparable number of resources.

\begin{table*}[htbp]
\vspace{-4mm}
\small
\addtolength{\tabcolsep}{-3pt}

\centering

\centering
\caption{\label{tab:vn_result} 
We report several metrics regarding the effectiveness and efficacy of \savag. First, we manage to mount very effective link prediction attacks. Second, thanks to our penalized framework, we are able to greatly reduce the amount of vicious nodes required. KL is in reported in base $10^{-3}.$
}

\begin{tabular}{lllllllllllll}

\toprule
                & \multicolumn{4}{c}{\textit{Twitter}} & \multicolumn{4}{c}{\textit{GPlus}} & \multicolumn{4}{c}{\textit{Citation2}}\\ \cmidrule(lr){2-5}\cmidrule(lr){6-9}\cmidrule(lr){10-13}
Method          & AR $\uparrow$ & AN $\downarrow$ & AP $\uparrow$ & KL $\downarrow$ & AR $\uparrow$ & AN $\downarrow$ & AP $\uparrow$ & KL $\downarrow$ & AR $\uparrow$ & AN $\downarrow$ & AP $\uparrow$ & KL $\downarrow$ \\ \midrule
\savag    & $0.75$ & $\mathbf{57.95}$  & $0.73$ & $3.00$ & $0.80$ & $\mathbf{35.70}$ & $0.79$ & $0.03$ & $0.85$ & $13.10$ & $0.87$ & $1.00$   \\ 
\savag-I    & $0.75$ & $61.05$  & $0.70$ & $3.00$ & $\mathbf{0.97}$ & $38.70$ & $\mathbf{0.95}$ & $0.03$ & $\mathbf{1.00}$ & $\mathbf{7.05}$  & $0.98$ & $0.30$   \\
\savag-N       & $0.95$ & $95.30$  & $0.93$ & $4.00$ & $0.93$ & $46.70$ & $0.90$ & $1.00$ & $0.85$ & $37.70$ & $0.89$ & $4.00$   \\    
\savag-NI     & $\mathbf{1.00}$ & $95.55$  & $\mathbf{0.99}$ & $4.00$ & $\mathbf{0.97}$ & $50.00$ & $\mathbf{0.95}$ & $0.03$ & $\mathbf{1.00}$ & $47.10$ & $\mathbf{0.99}$ & $3.00$   \\ 
AIGA      & $0.60$ & $100.0$ & $0.61$ & $\mathbf{0.00}$             & $0.23$ & $50.00$ & $0.29$ & $\mathbf{0.00}$             & $0.90$ & $50.00$ & $0.93$ & $\mathbf{0.03}$ \\ \midrule
 RAND-L  & $0.00$ & $23.15$  & $0.04$ & $5.00$ & $0.00$ & $13.70$ & $0.04$ & $1.00$ &  $0.05$      & $12.95$        & $0.03$        & $6.00$ \\ 
RAND-H  & $0.10$ & $74.70$  & $0.11$ & $10.00$ & $0.00$ & $37.35$ & $0.06$ & $10.00$ &  $0.00$      & $38.85$         & $0.03$        & $10.00$ \\ \midrule 
\end{tabular}

\vspace{1mm}

\begin{tabular}{lllllllllllllllll}
\toprule
                & \multicolumn{4}{c}{\textit{Arxiv}} & \multicolumn{4}{c}{\textit{Cora}} & \multicolumn{4}{c}{\textit{Wiki}} & \multicolumn{4}{c}{\textit{Synthetic}}\\ \cmidrule(lr){2-5}\cmidrule(lr){6-9}\cmidrule(lr){10-13}\cmidrule(lr){14-17}
Method          & AR $\uparrow$ & AN $\downarrow$ & AP $\uparrow$ & KL $\downarrow$ & AR $\uparrow$ & AN $\downarrow$ & AP $\uparrow$ & KL $\downarrow$ & AR $\uparrow$ & AN $\downarrow$ & AP $\uparrow$ & KL $\downarrow$ & AR $\uparrow$ & AN $\downarrow$ & AP $\uparrow$ & KL $\downarrow$ \\ \midrule
\savag     & $0.80$ & $35.70$ & $0.80$ & $5.00$ & $0.83$ & $47.60$ & $0.83$ & $1.00$ & $\mathbf{1.00}$ & $14.20$ & $0.88$ & $7.00$ & $\mathbf{1.00}$ & $14.00$ & $0.88$ & $7.00$   \\ 
\savag-I  & $0.75$ & $\mathbf{34.90}$ & $0.75$ & $5.00$ & $\mathbf{0.96}$ & $\mathbf{33.60}$ & $\mathbf{0.97}$ & $\mathbf{0.04}$ & $0.60$ & $\mathbf{10.25}$ & $0.63$ & $3.00$ & $0.60$ & $\mathbf{9.00}$ & $0.63$ & $3.00$\\
\savag-N       & $\mathbf{0.95}$ & $44.55$ & $\mathbf{0.93}$ & $5.00$ & $0.83$ & $59.80$ & $0.83$ & $3.00$ & $\mathbf{1.00}$ & $29.70$ & $\mathbf{0.96}$ & $7.00$ & $\mathbf{1.00}$ & $29.00$ & $\mathbf{0.96}$ & $7.00$   \\    
\savag-NI     & $0.90$ & $43.05$ & $0.89$ & $5.00$ & $0.70$ & $49.40$ & $0.69$ & $3.00$ & $0.70$ & $11.10$ & $0.68$ & $3.00$ & $0.70$ & $11.00$ & $0.68$ & $3.00$   \\ 
AIGA      & $0.70$ & $50.00$ & $0.67$ & $\mathbf{0.00}$                & $0.80$ & $80.00$   & $0.79$ & $0.40$ & $0.40$ & $50.00$   & $0.47$ & $\mathbf{0.00}$                & $0.90$ & $50.00$ & $0.93$ & $\mathbf{0.03}$   \\ \midrule
RAND-L  & $0.10$ & $11.40$ & $0.11$ & $6.00$ & $0.00$ & $17.50$         & $0.00$     &    $4.00$              & $0.10$     & $13.10$     & $0.29$      & $3.00$ & $0.10$    & $13.10$ & $0.31$ & $6.00$ \\ 
RAND-H  & $0.15$ & $37.70$ & $0.14$ & $10.00$ & $0.00$     & $60.15$     & $0.00$     & $10.00$  & $0.05$     & $36.65$     & $0.29$      & $10.00$  & $0.10$    & $37.40$   & $0.22$ &$30.00$ \\ \bottomrule
\end{tabular}

\end{table*}

\subsection{Ablation Study}

We perform an extensive ablation study to support the design choices made for the main components of our framework.
A key characteristic of \savag\ is its ability to enforce sparsity of malicious resources by adjusting the optimization goal of the attacker.
We analyze several components of the model, like the initialization of the perturbation matrix and the effect of the penalty losses, to study their behavior and how this impacts the sparsification effort.

\begin{figure}
    \centering
    \includegraphics[scale=0.16]{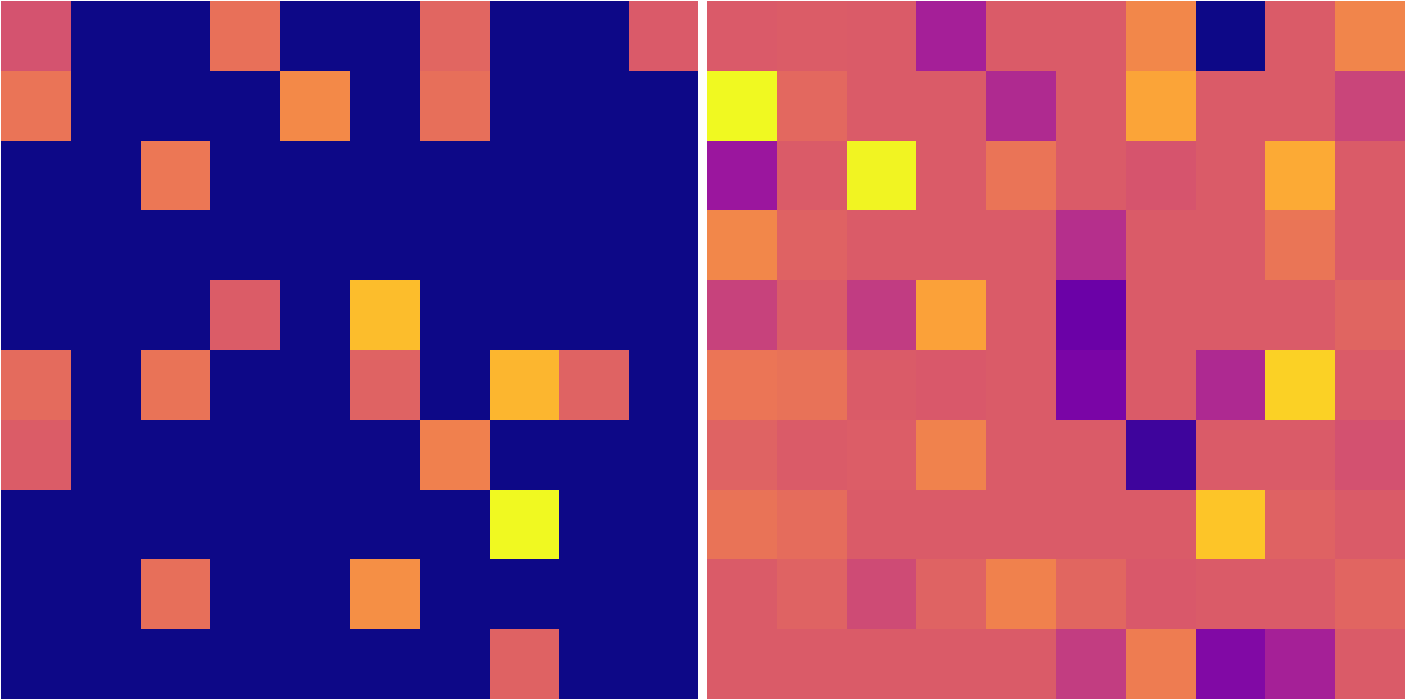}
    \caption{Visualizations of gradients for the model with $\ldist$ and $\lnew$ (left), and without (right). Each cell describes the activation strength of a vicious node. In blue are deactivated nodes. We (visually) show that our losses successfully induce a sparse usage of malicious resources.}
    \label{fig:sparse_matrix}
\end{figure}

We provide here a graphic visualization of our findings on \savag\ inner working.
In Fig.~\ref{fig:sparse_matrix}, we inspect the gradients of our model with our introduced sparsity losses $\ldist$ and $\lnew$ (left) against the model not featuring them (right).
It is highly evident that most vicious nodes on the left are deactivated (encoded with blue), and do not contribute as additional resources to mount the attack.
This is not true for gradients on the right, which instead have a much higher diffusion, with credits being awarded to the largest number of vicious nodes possible.
Thus, we argue this is further proof that our formulation is effective in injecting sparsity into the model, making more efficient usage of malicious resources.
Again, we remind that we do not directly optimize for the feature matrix of the graph, focusing instead on its topology, and leaving it for future work.
Still, we conduct a comprehensive analysis, showing that \savag\ is robust under different assumptions for the creation of the vicious nodes' features.
For the entire set of results from our ablation study we refer the reader to the supplementary material.



\subsection{Attack Transferability}
\label{subsec:transf}

\savag\ defines a white-box attack model, i.e., it assumes a full knowledge of the target GNN-based link prediction model. 
This assumption may be unrealistic in practice; indeed, the internals of a link prediction system (e.g., the model's architecture, parameters, or gradients) are rarely disclosed, especially in a commercial scenario.
Nevertheless, we claim that this restriction does not limit the feasibility of our method. 
In fact, we show that the adversarial perturbations generated by \savag\ can be successfully transferred to fool other link prediction systems (not necessarily GNN-based) under the more challenging black-box setting.
Specifically, to demonstrate this capability, we focus on several link prediction heuristics, as reported in Table~\ref{tab:tra_alg}.
The higher the value of these heuristics, the higher the chance of a link between two nodes. 
For a given graph, a source and a target node initially disconnected, and a white-box GNN-based link predictor, we generate the corresponding perturbation with \savag\ so to increase the probability of a connection between the victim and the source. 
The scores in Table~\ref{tab:tra_alg} are expressed in terms of the lift of the heuristic values as computed in the original, unperturbed graph, i.e., before and after the attack, averaged across 20 different pairs of sampled source and target nodes.
We observe an increase of all the heuristic values considered, which indicates that \savag\ would be able to mount attacks even on non-GNN-based link prediction systems.


\begin{table}[htbp]
\centering
\small
\addtolength{\tabcolsep}{-8pt}
\caption{\label{tab:tra_alg} Lift scores for popular link prediction heuristics before and after being attacked by \savag. 
Given the large increase in heuristic values after the attack, we claim that our attacks successfully transfer to different black-box methods.
}
    \begin{tabular}{cc}
    \toprule
    Heuristic & \begin{tabular}[c]{@{}c@{}}Lift Score \\ (attack/non-attack)\end{tabular} \\\midrule
    Common Neighbors~\cite{newman2001clustering} & $70 \times$\\
    Jaccard~\cite{liben2003cikm} & $93\times$\\
    Preference Attachment~\cite{barabasi1999pa} & $3.4\times$\\
    Adam-Adar~\cite{adamic2003friends} & $61\times$\\
    Resource Allocation~\cite{zhou2009predicting} & $2.9\times$\\
    Katz Index~\cite{katz1953new} & $3 e5\times$\\
    PageRank~\cite{brin1998pagerank} &  $1.2\times$\\
    SimRank~\cite{jeh2002simrank} & $20\times$\\
    \bottomrule
    \end{tabular}

\end{table}


\subsection{Limitations}
\label{subsec:limit}
Experiments have demonstrated that \savag\ can achieve a high attack success rate while using a sparse number of new, malicious nodes.
However, \savag\ causes a shift in the node's degree distribution between the original and the attacked graph larger than other methods, thus indicating a more substantial perturbation of the initial graph links.
We argue that, in general, from the attacker's perspective, the cost of adding a vicious node to the network is the most relevant and that should receive the highest priority.
Moreover, on average, the KL-divergence between the two node distributions is smaller than $10^{-3}$ with \savag. Still, we acknowledge this as a limitation of our current method.
We plan to incorporate a new factor (based on the KL-divergence above) in the attacker's objective, limiting the impact of \savag\ on the original graph connections. 

\section{Conclusion and Future Work}
\label{sec:conclusion}
In this work, we presented \savag, a novel framework for generating adversarial attacks on GNN-based link prediction systems.
\savag\ defines a white-box attack model where an adversary aims to appear in the list of recommended users to follow for a target victim node.
The attacker controls a subset of original nodes and some additional vicious nodes that can inject into the network. Each of these malicious nodes can, in turn, create or destroy direct edges with others until eventually altering the link prediction between the target victim node and the attacker node.
\savag\ formulated the problem that the attacker must solve as an optimization task, which trades off between the attack's success and the sparsity of malicious resources required.
Experiments conducted on real-world and synthetic datasets showed the effectiveness and efficiency of our approach in attacking several GNN-based link prediction models. Moreover, we showed that \savag\ attacks can be successfully transferred to disrupt other link prediction systems (not necessarily GNN-based) under black-box setting.
In future work, we plan to extend our attack model with the ability to perturb node features. 
Furthermore, our method must operate on dense adjacency matrices, somewhat limiting its scalability. 
Thus, we will develop a variant of \savag\ that can work with sparse adjacency matrices.
The design of possible defense mechanisms to combat the attacks generated by \savag\ is another interesting direction to explore; we will also consider different downstream tasks as the target of the attacks (e.g., node classification).


\clearpage

\bibliography{main.bib}

\clearpage

\appendix

\section{Datasets}

\paragraph{Pruning}
Given both the high number of nodes in the social network-based datasets (Twitter and GPlus) and the high dimensionality of their feature spaces, we decided to prune them.
We started by reducing the noise in the Twitter features by normalizing the textual versions to lowercase and removing characters not allowed in the hashtag and mention as per Twitter official guidelines.\footnote{https://help.twitter.com/en/managing-your-account/twitter-username-rules\#error} Then, for Twitter, we removed features with less than 20 occurrences, and for GPlus, the ones with less than 10. After that, we pruned users with less than 20 remaining features on Twitter and 30 for GPlus. We report in Table~\ref{tab:pruning} the pruning results.

\paragraph{Preprocessing}
Considering the high-dimensionality of the majority of the dataset used and their sparsity, we adopted a preprocessing to make them smaller and, simultaneously, as denser as possible. In particular, we extracted the largest connected components from all of them. Subsequently, if the extracted subgraph had more than 10K nodes, we first sampled employing GraphSAINTSampler \cite{Zeng2020GraphSAINT}, then we retrieved the largest connected component again to ensure the density of the sampled subgraph. We report in Table~\ref{tab:dat_stat} the statistics of the preprocessed dataset.

\begin{table}[htbp]
\caption{\label{tab:pruning} User and feature pruning results for each social network dataset. We also consider the shrink percentage between the starting value and the one after the pruning.}
\centering
\begin{tabular}{ccrr}
\toprule
 &  & GPlus & Twitter \\
 \midrule
\multirow{3}{*}{\rotatebox[origin=c]{90}{Features}} & Original & 15,602 & 155,033 \\
 & Pruned & 2,001 & 11,034 \\
 & Shrink & 88\% & 93\% \\
 \midrule
\multirow{3}{*}{\rotatebox[origin=c]{90}{Users}} & Original & 107,613 & 81,306 \\
 & Pruned & 7,008 & 7,520 \\
 & Shrink & 93\% & 91\%\\ \bottomrule
\end{tabular}
\end{table}
\begin{table}[htbp]
\centering
\caption{\label{tab:dat_stat}We report the statistics of the datasets used. In particular, $N$ represents the number of nodes, $E$ represents the number of edges, $ \bar{X} $ represents the mean between the in and the out degree, $ \Tilde{X}$ represents the median between the in and the out degree and $AC$ represents the average clustering.}
\begin{tabular}{llllll}
\toprule
Dataset   & \multicolumn{1}{c}{ $N$ } & \multicolumn{1}{c}{ $E$ } & \multicolumn{1}{c}{$ \bar{X} $} & \multicolumn{1}{c}{$ \Tilde{X} $} & \multicolumn{1}{c}{\shortstack{$AC$}} \\ \midrule
Twitter   & $7.50$K & $725.20$K & $96.84$ & $28.50$ & $0.43$ \\
Gplus     & $6.90$K & $2.80$M  & $403.20$ & $35.00$ & $0.28$ \\
Citation2 & $10.10$K & $15.40$K & $1.52$ & $1.00$  & $0.07$ \\
Arxiv     & $10.70$K & $46.60$K & $4.36$ & $2.00$  & $0.13$ \\
Cora      & $2.50$K & $5.20$K  & $2.10$  & $1.50$ & $0.13$ \\
Wiki      & $7.00$K & $103.70$K & $14.67$ & $2.00$  & $0.08$ \\
Synthetic & $5.50$K & $54.70$K & $9.99$ & $10.00$ & $0.01$ \\ \bottomrule
\end{tabular}
\end{table}

\section{GNN Models}

We train our link prediction system in a transductive setting, using $90\%$ of existing edges as training set and the remaining $10\%$ as test set.
We train our models for $2000$ epochs with a learning rate $lr=0.001$.
The performance of our GNN-based link prediction models measured on the test set for each dataset considered is reported in Table~\ref{tab:auc}.

\begin{table}[htbp]
\centering
\caption{\label{tab:auc}We report Accuracy and AUROC scores obtained by our link prediction systems on the datasets used in our experiments. Given the high scores obtained, we are confident to attack a strong-enough prediction system.}
\begin{tabular}{@{}lcc@{}}
\toprule
Dataset   & \multicolumn{1}{l}{Accuracy} & \multicolumn{1}{l}{AUROC} \\ \midrule
Twitter   & $0.94$                         & $0.94$                      \\
GPlus     & $0.81$                         & $0.81$                      \\
Citation2 & $0.97$                         & $0.97$                      \\
Arxiv     & $0.90$                         & $0.89$                      \\
Cora      & $0.82$                         & $0.82$                      \\
Wiki      & $0.78$                         & $0.78$                      \\
Synthetic & $0.77$                         & $0.77$                      \\ \bottomrule
\end{tabular}
\end{table}

\section{Methods}

We have reported results for different configurations of our framework. In particular, our SAVAGE and SAVAGE-I configurations rely on a set of hyperparameters: the penalty losses $\beta$ and $\gamma$. 
In Table~\ref{tab:hyperParam} we report the best hyperparameter setting. 
These hyperparameters were found through a grid search in the range $[10^{-3}, 10]$.

\begin{table}[htbp]
\centering
\caption{\label{tab:hyperParam} Best hyperparameter settings for selected configuration and dataset.}
\begin{tabular}{@{}lcccc@{}}
\toprule
          & \multicolumn{2}{c}{SAVAGE}                   & \multicolumn{2}{c}{SAVAGE-I}                 \\ \cmidrule(l){2-5} 
Dataset   & $\beta$ & $\gamma$ & $\beta$ & $\gamma$ \\ \midrule
Twitter   & $0.80$                    & $0.80$                   & $0.80$                  & $0.80$                   \\
GPlus     & $0.80$                  & $0.80$                   & $0.80$                  & $0.80$                  \\
Citation2 & $0.80$                  & $0.80$                   & $0.80$                  & $0.80$                   \\
Arxiv     & $0.03$                 & $0.03$                  & $0.03$                 & $0.03$                  \\
Cora      & $0.80$                  & $0.80$                   & $0.80$                  & $0.80$                   \\
Wiki      & $0.10$                  & $0.10$                   & $0.01$                 & $0.01$                  \\
Synthetic & $0.10$                  & $0.10$                  & $0.01$                 & $0.01$                  \\ \bottomrule
\end{tabular}
\end{table}

\section{Ablations}

\subsection{Initialization}

As hinted several times in the paper, we are solving a highly non-convex optimization problem.
For this reason, the initialization of the perturbation matrix $P$ can be considered crucial, as it reflects both the prior information we inject into the model and the model's results itself.
For this reason, we run an ablation experiment to show how different kinds of initializations lead to different results.

We test four possible initializations for the perturbation matrix: Random, All Zeros + $\epsilon$, all Ones - $\epsilon$, and All negative ones + $\epsilon$; where $\epsilon$ is a small random positive number ($< 0.3$).
We test these options by running experiments for $20$ different pairs randomly sampled from the Synthetic dataset.

We report results in Table~\ref{tab:init_P}. We indicate the All negative ones option with a minus as footer (Ones\_).

We may observe that Random exhibits the best trade-off between the attack power and the number of resources used.

\begin{table}[htbp]
\centering
\caption{\label{tab:init_P} Results for different initialization of the perturbation matrix $P$. The Random initialization offers the best trade off between resources used and power of the attack.}
\begin{tabular}{@{}lcccc@{}}
\toprule
Method      & AR   & AN    & AP   & KL \\ \midrule
Random      & $0.90$ & $14.00$  & $0.89$ & $0.01$  \\
Zeros       & $0.75$ & $11.50$  & $0.74$ & $0.01$  \\
Ones        & $0.45$ & $7.20$   & $0.50$ & $0.01$  \\
Ones\_ & $0.55$ & $22.05$ & $0.58$ & $0.00$  \\ \bottomrule
\end{tabular}
\end{table}

\subsection{Penalty losses}

SAVAGE relies on two penalty losses, $\beta$ and $\gamma$, to effectively sparsify the resources used to carry out the attack.
We hereby perform a sensitivity analysis to quantify the impact of these losses on the model results.
In particular, we sample $20$ pairs at random from the Synthetic dataset and run our method on them.

\paragraph{L-Dist}

\begin{figure}[htb]
\centering
    \begin{tikzpicture}
\begin{axis}[
    title={Ablation Study on $\beta$},
    xlabel={$\beta$},
    ylabel={Nodes Used},
    xmin=0, xmax=10,
    ymin=0, ymax=15,
    xtick={},
    ytick={},
    legend pos=north west,
    ymajorgrids=true,
    grid style=dashed,
]
\addplot[
    color=blue,
    mark=square,
    ]
    coordinates {
    (0, 14.85)
    (1, 7.6)
    (2, 5)
    (3, 3.5)
    (4, 2.95)
    (5, 2.6)
    (6, 2.15)
    (7, 2)
    (8, 1.7)
    (9, 1.55)
    (10, 1.5)
    };
\end{axis}
\end{tikzpicture}%
\caption{Amount of resources used at the variying of the penalty loss $\beta$. It can be noticed how the decrease in the resources used produces a well behaved monotonic curve.}
    
    \label{fig:abl_betas}
\end{figure}
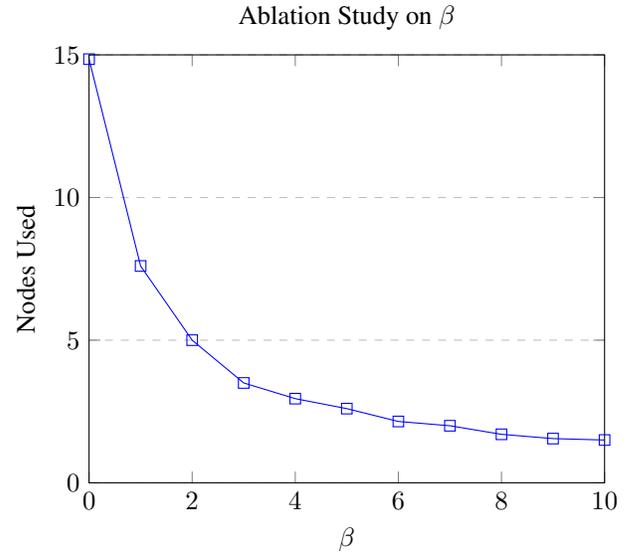

As explained in the relevant section of the main body, $\beta$ regulates the effect of $\ldist$, that is, the penalty discouraging $\tilde{h}_{s;P}$ from being too far off from $h_s$.
In Figure~\ref{fig:abl_betas} we report our finding. On the horizontal axis, we indicate the penalty used, while on the vertical axis, the amount of resources used.
As we can see, increasing $\beta$ corresponds to a reduction in the resources used.
Furthermore, we notice that the behavior of this penalty loss is optimal, as it features a nice monotonic decrease in the amount of resources used.
While not shown here for visualization purposes, the relationships remain pretty much constant outside the range considered.

\paragraph{L-new}

\begin{figure}[ht!]
\centering
    \begin{tikzpicture}
\begin{axis}[
    title={Ablation Study on $\gamma$},
    xlabel={$\gamma$},
    ylabel={log(Nodes Used)},
    xmin=0, xmax=10,
    ymin=-2.5, ymax=4,
    xtick={},
    ytick={},
    legend pos=north west,
    ymajorgrids=true,
    grid style=dashed,
]
\addplot[
    color=blue,
    mark=square,
    ]
    coordinates {
    (0, 3.18)
    (1, 0.97)
    (2, 0.44)
    (3, 0.05)
    (4, 0.05)
    (5, -0.51)
    (6, -0.6)
    (7, -0.7)
    (8, -1.6)
    (9, -1.05)
    (10, -1.9)
    };
\end{axis}
\end{tikzpicture}%
\caption{Amount of resources used (log-scale) at the variying of the penalty loss $\gamma$. Again, it can be noticed how the decrease in the resources used produces a well behaved, almost monotonic, curve.}
    \label{fig:abl_gammas}
\end{figure}
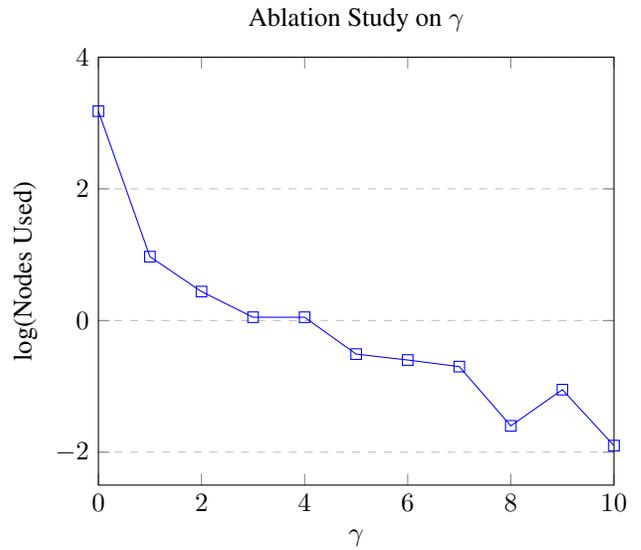

As explained in the relevant section of the main body, $\gamma$ regulates the effect of $\lnew$, that is, the penalty directly discouraging the proliferation of vicious nodes injected into the original graph network.
We report results in Figure~\ref{fig:abl_gammas}.
We indicate the level of the penalty $\gamma$ used on the horizontal axis, with the amount of resources used on the vertical axis (on the log scale).
Once again, even though the intrinsic discreteness of the problem, we show how our formulation features a nice, well-behaved decrease in resources used along with the increase in the penalty loss, with an almost-monotonic curve.
While not shown here for visualization purposes, the relationships remain pretty much constant outside the range considered.

\subsection{Feature Matrix}

As stated in the main body of this paper, SAVAGE does not focus on the creation of the features for the vicious nodes injected into the original graphs, focusing instead on their topology.

\begin{table}[htbp]
\centering
\caption{\label{tab:X}Results under different assumptions for creating the vicious nodes feature matrix. While not optimizing for this, SAVAGE is robust under different settings and maintains strong results.}
\begin{tabular}{@{}lcccc@{}}
\toprule
Method   & AR   & AN   & AP   & KL   \\ \midrule
Existent & $0.90$  & $13.90$ & $0.90$  & $0.01$ \\
Random   & $1.00$    & $16.70$ & $0.98$ & $0.01$ \\
Ones     & $0.75$ & $42.50$ & $0.75$ & $0.01$ \\
Zeros    & $0.95$ & $14.10$ & $0.91$ & $0.01$ \\
Mean     & $0.91$ & $14.10$ & $0.91$ & $0.01$ \\
Median   & $0.95$ & $14.00$ & $0.91$ & $0.01$ \\ \bottomrule
\end{tabular}
\end{table}

In particular, we do not optimize for it, leaving it as a feature work.
However, we hereby claim that SAVAGE 
is robust under different assumptions for the creation of such a matrix.
For this experiment, once again, we sampled $20$ pairs randomly from the Synthetic dataset and ran our method on them.
This time, however, we consider six different settings for creating the feature matrix.
The first setting, used for the main experiment, is to randomly sample features from existing nodes and add a slight noise.
Three settings consist in initializing the feature matrix $X$ with zeros, ones, and at random, respectively; we apply some noise $\epsilon$ here as well.
Finally, we take the mean and the median of the feature matrix in two cases, still adding noise to it.
Results are reported in Table~\ref{tab:X}. 
These results show that we maintain strong results across different initializations.
In particular, we can notice how the Random, Zeros, and Median methods produce the strongest attack power, while Ones produces the lowest attack power while using the most resources as well.

\end{document}